\documentclass{article}

     \usepackage[preprint,nonatbib]{neurips_2020}

\usepackage[utf8]{inputenc} 
\usepackage[T1]{fontenc}    
\usepackage{hyperref}       
\usepackage{url}            
\usepackage{booktabs}       
\usepackage{amsfonts}       
\usepackage{nicefrac}       
\usepackage{microtype}      

\usepackage{times}
\usepackage{epsfig}
\usepackage{graphicx}
\usepackage{amsmath}
\usepackage{bbm, dsfont}
\usepackage{amsthm}
\usepackage{amssymb}
\usepackage{booktabs}
\usepackage[dvipsnames, table]{xcolor}
\definecolor{Gray}{gray}{0.85}
\usepackage{multirow}
\usepackage{pifont}
\newcommand{\cmark}{\textcolor{OliveGreen}{\ding{51}}}%
\newcommand{\xmark}{\textcolor{red}{\ding{55}}}%
\usepackage{algorithm}
\usepackage{algpseudocode}
\usepackage{comment}
\usepackage{hyperref}
\usepackage{subcaption}
\usepackage{caption}
\usepackage{tikz}
\usepackage{comment}
\usepackage{amsmath,amssymb} 
\usepackage{color}

\title{SPIQ: Data-Free Per-Channel Static Input Quantization}

\author{%
  Edouard Yvinec$^{1,2}$ \And Arnaud Dapogny$^2$ \And Matthieu Cord$^{2,3}$ \And Kevin Bailly$^{1,2}$  \\
  $^1$ Sorbonne Université, CNRS, ISIR, f-75005, 4 Place Jussieu 75005 Paris, France\\
  $^2$ Datakalab, 114 boulevard Malesherbes, 75017 Paris, France \\
  $^3$ Valeo, 100 rue de Courcelles, 75017 Paris, France \\
  \texttt{ey@datakalab.com}\\
}

\begin{document}

\maketitle
\begin{abstract}
Computationally expensive neural networks are ubiquitous in computer vision and solutions for efficient inference have drawn a growing attention in the machine learning community. Examples of such solutions comprise quantization, \textit{i.e.} converting the processing values (weights and inputs) from floating point into integers e.g. int8 or int4. Concurrently, the rise of privacy concerns motivated the study of less invasive acceleration methods, such as data-free quantization of pre-trained models weights and activations. Previous approaches either exploit statistical information to deduce scalar ranges and scaling factors for the activations in a static manner, or dynamically adapt this range on-the-fly for each input of each layers (also referred to as activations): the latter generally being more accurate at the expanse of significantly slower inference. In this work, we argue that static input quantization can reach the accuracy levels of dynamic methods by means of a per-channel input quantization scheme that allows one to more finely preserve cross-channel dynamics. We show through a thorough empirical evaluation on multiple computer vision problems (e.g. ImageNet classification, Pascal VOC object detection as well as CityScapes semantic segmentation) that the proposed method, dubbed SPIQ, achieves accuracies rivalling dynamic approaches with static-level inference speed, significantly outperforming state-of-the-art quantization methods on every benchmark.

\end{abstract}
\section{Introduction}
Deployment of State-of-the-art deep neural networks (DNNs) on edge devices has become increasingly difficult. Although edge computing has recently drawn more attention, motivated by privacy \cite{shi2016promise} and environmental sustainability concerns \cite{li2018green}, concurrently DNNs have grown more computationally expensive. Fortunately, DNNs parameters encompass abundant opportunities for trimming strategies such as quantization. 

As defined in \cite{gholami2021survey}, quantization consists in mapping a set of continuous variables to a finite set of values, e.g. int8, int4 or ternary, in order to compress the bit-wise representation. Quantization trends can be distinguished by data-usage and range calibration. 

First, the approximation introduced by quantization often requires adjustments in order to preserve the original accuracy of the model. This can be performed using real training data and  is called data-driven quantization \cite{gysel2018ristretto,lin2015neural,tailor2020degree,gysel2016hardware}. Although such methods can afford lower bit-wise representations, they are both computationally expensive and less convenient to use. On the other hand, when quantization is performed without re-training, it is often referred to as post-training quantization (PTQ) or data-free quantization \cite{cai2020zeroq,fang2020post,zhao2019improving,nagel2019data,squant2022}. Such methods are convenient for applications where privacy and security are mandatory. The proposed method aims at reducing the gap between data-free and data-driven quantization by improving input and activation quantization.

Second, in order to quantize inputs, the range of their distribution has to be estimated.
In data-free quantization, inputs of each layer are quantized based on statistics determined either from the already trained parameters (static quantization) \cite{nagel2019data} or based on statistics computed on-the-fly based on each sample at inference (dynamic quantization) \cite{sun2021dynamic}. The latter usually offers significantly higher accuracy at the expense of a slower inference, more-so on low bit-wise representations. 

\begin{figure}[t]
    \centering
    \includegraphics[width = \linewidth]{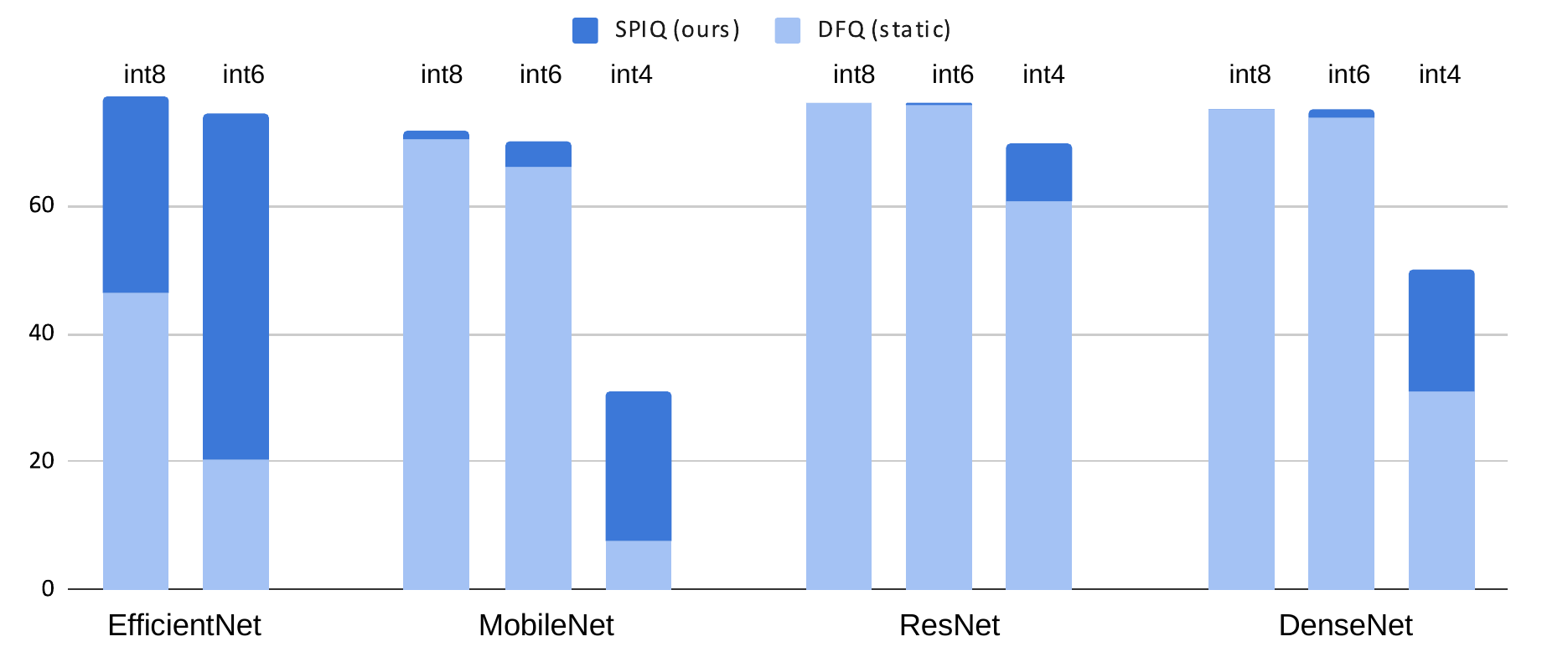}
    \caption{Illustration of the accuracy drop attributable to the input and activation quantization. We perform input quantization as defined in \cite{nagel2019data} as well as SPIQ (ours) but keep the original weight values, \textit{i.e.} only the input of each layer is quantized. The results show that input quantization is paramount to the network accuracy preservation, most notably on already compact designs (e.g. MobileNet and EfficientNet). On all tested configurations, SPIQ significantly outperforms DFQ \cite{nagel2019data}.}
    \label{fig:related}
\end{figure}

While most research on data-free quantization \cite{banner2019post,cai2020zeroq,choukroun2019low,fang2020post,garg2021confounding,zhao2019improving,nagel2019data} focuses on weight quantization, we provide empirical evidence that input quantization is responsible for a significant part of the accuracy loss, most notably on low bit representation, as illustrated in Fig \ref{fig:related}. Furthermore, we show that per-channel input range estimation allows tighter modelling of the full-precision distribution as compared to a per-example, dynamic approach. As a result, the proposed SPIQ (standing for \textbf{S}tatic \textbf{P}er-channel \textbf{I}nput \textbf{Q}uantization) method outperforms both static and dynamic approaches as well as existing state-of-the-art methods.
\section{Related Work}

\subsection{Quantization}
As pointed out in \cite{gray1998quantization} by Gray and Neuhoff, quantization as a compression method transforming continuous values to discreet ones, has a long history. Rounding and truncating are the most common examples. As discussed in \cite{gholami2021survey}, quantization methods are classified as either data-driven \cite{gysel2018ristretto,hubara2016binarized,lin2015neural,tailor2020degree,courbariaux2015binaryconnect,gysel2016hardware} or data-free \cite{banner2019post,cai2020zeroq,choukroun2019low,fang2020post,garg2021confounding,zhao2019improving,nagel2019data,squant2022}. Data-driven methods have been shown to work remarkably well despite a coarse approximation of the continuous optimisation problem. However, the cost of retraining the model has limited the use of such solutions in inference engine and in general for Machine Learning as a Service (MLaaS) \cite{ribeiro2015mlaas}. Furthermore, with the rise of privacy concerns especially in health services \cite{zhang2018health}, data-free methods are becoming of most importance. However such methods often come at the cost of a lower accuracy.

\subsection{Data-Free Quantization}
The importance of data-free quantization is discussed in more details, in the most recent survey on the matter \cite{gholami2021survey}.
Most data-free methods focus on mitigating the accuracy drop resulting from the quantization process. For instance, DFQ \cite{nagel2019data} proposes to balance-out weight distribution across layers in order to reduce the bias induced by quantization. They also propose the first static input quantization method based on learned statistics stored in batch-normalization layers parameters. In SQuant \cite{squant2022}, authors propose to further improve weight quantization by changing the implicit objective function. More formally, rounding scalar weights minimizes the mean squared error between the scalar quantized weights and the original weights. SQuant minimizes the absolute sum of errors between tensor instead of scalar values. Similarly most data-free methods \cite{banner2019post,cai2020zeroq,choukroun2019low,fang2020post,garg2021confounding,zhao2019improving,nagel2019data} focus solely on improving weight quantization.

\subsection{Input Quantization}
In this work, we put the emphasis on the importance of input quantization, especially for the already compact architectures such as MobileNet \cite{sandler2018MobileNetV2} and EfficientNet \cite{tan2019efficientnet} as previously shown in Fig \ref{fig:related}. The standard method for input quantization, introduced in \cite{nagel2019data} and latter used in \cite{squant2022,zhang2021diversifying}, is static as its parameters are computed once and for all during quantization then fixed during inference. This provides the best inference speed but for at the cost of a lower accuracy due to coarse modelization of the input ranges. To mitigate the accuracy drop, dynamic input quantization \cite{sun2021dynamic} consists in computing the quantization scale per inference sample, improving the accuracy by tight adaptation to the input statistics at the cost of inference speed. The other solution consists in dropping the data-free constraint and fitting the input quantization parameters on a validation sample \cite{zhao2019improving,choukroun2019low,shin2016fixed,sung2015resiliency,park2017weighted} or during an extensive training \cite{bhalgat2020lsq+,esser2019learned,choi2018pact}.

In this work, we provide a solution for static input quantization that reaches accuracy levels of dynamic methods by means of a per-channel input quantization scheme that allows one to more finely preserve cross-channel dynamics
\section{Methodology}
Let $F : \mathcal{D}\mapsto \mathbb{R}^{n_o}$ be a feed forward neural network defined over a domain $\mathcal{D}\subset \mathbb{R}^{n_i}$ and output space $\mathbb{R}^{n_o}$. The operation performed by a layer $f_l$, for $l \in \{1,\dots,L\}$, is defined by the corresponding weight tensor $W_l\in \mathcal{A}^{n_{l-1}\times n_l}$ where $\mathcal{A}$ is simply $\mathbb{R}$ in the case of fully-connected layers and $\mathbb{R}^{k\times k}$ in the case of a $k\times k$ convolutional layer. We note $I_l$ the input of a fully-connected layer $f_l$. Let's consider a quantization operator $Q:\mathbb{R}\rightarrow [-\beta;\beta] \cap \mathbb{N}$ which maps real values to a bounded set of integer values where $\beta = 2^{b-1}-1$ and $b$ defines the bit-width of the target representation. The standard quantization operator is defined as $Q : x \mapsto \lfloor x / s_x \rceil$ where $\lfloor \cdot \rceil$ is the rounding operation and $s_x$ is a scaling factor. Then, the quantized layer $f^q_l$ is defined as
\begin{equation}\label{eq:def}
    f_l^q : I_l \mapsto Q^{-1} \left( Q(I_l) \times Q(W_l)\right) = s_{I_l} \odot s_{W_l}\odot \left( \left\lfloor \frac{I_l}{s_{I_l}} \right\rceil \times \left\lfloor \frac{W_l}{s_{W_l}} \right\rceil\right)
\end{equation}
where $\odot$ is the element-wise product.
The values of $s_{I_l}$ and $s_{W_l}$ depend on the information available on $I_l$ and $W_l$ respectively. In the case of weight tensor $W_l$ during the quantization process, all the information is available. Consequently, the value of $s_{W_l}$ is derived from $W_l$ in order to scale the scalar weight values distribution to $[-\beta;\beta]$. There are two quantization options. First, per output-channel weight quantization, in this case $s_{W_l}\in\mathbb{R}_+^{n_l}$ is a $n^l-$dimensional vector and each output channel (or neuron) is scaled independently. Second, per-layer (or per-tensor) quantization, where $s_{W_l}\in\mathbb{R}_+$ is a scalar value that scales the whole weight tensor $W_l$. Formally, if the note $W_l^{\text{channel}}$ the per-channel quantized tensor and $W_l^{\text{layer}}$ the per-layer quantized tensor,
\begin{equation}
\begin{cases}
     W_l^{\text{channel}} &= \left\lfloor (2^{b-1}-1)\frac{W_l}{(\max_{w\in W_l^n}\{|w|\})_{n\in\{1,\dots,n_l\}}} \right\rceil\\
     W_l^{\text{layer}}   &=\left\lfloor (2^{b-1}-1)\frac{W_l}{\max_{w\in W_l}\{|w|\}} \right\rceil
\end{cases}
\end{equation}
where $W_l^n$ is the $n^{\text{th}}$ column of $W_l$ corresponding to the $n^{\text{th}}$ neuron of layer $f_l$.
\subsection{Static and Dynamic Input Quantization}\label{sec:3_1}
The definition of $s_{I_l}$ from equation \ref{eq:def} induces a dimensionality constraint. We need to apply $s_{I_l}$ to both $I_l$ (which has $n_{l-1}$ channels) and $\left\lfloor \frac{I_l}{s_{I_l}} \right\rceil \times \left\lfloor \frac{W_l}{s_{W_l}} \right\rceil$, \textit{i.e.} if $s_{I_l}$ is a vector then $s_{I_l}$ needs to be of dimension $n_{l-1}$ and $n_l$ in order to be applied to $I_l$ and $\left\lfloor \frac{I_l}{s_{I_l}} \right\rceil \times \left\lfloor \frac{W_l}{s_{W_l}} \right\rceil$ respectively. Therefore, $s_{I_l}$ has to be a single, scalar value that scales the whole input tensor.

Similarly to the weight scaling factor $s_{W_l}$, the input scale $s_{I_l}$ is computed based on the support of the distribution to scale. However, in the case of data-free quantization, we don't have access to the statistical properties of the input domain $\mathcal{D}$ of $F$. In order to circumvent this limitation we can apply either a static or dynamic activation quantization scheme.
\paragraph{Static Input Quantization:} The goal is to compute $s_{I_l}^{\text{static}}\in\mathbb{R}$ based on an estimation of the maximum of $I_l$ over the domain $\mathcal{D}$. Assume a BN layer precedes $f_l$, we can assert that
\begin{equation}
    \mathbb{E}[I_l]^n = \beta^n \quad \text{and}\quad \mathbb{V}[I_l]^n = \gamma^n
\end{equation}
where $\beta\in\mathbb{R}^{n_{l-1}}$ and $\gamma\in\mathbb{R}^{n_{l-1}}$ are the centering and scaling vector parameters of the BN layer respectively. Consequently, the maximum value of $I_l$ over the domain $\mathcal{D}$, can be derived by searching for the maximum over the output channels and we get,
\begin{equation}
     \frac{\max_{i\in I_l \text{ from }\mathcal{D}} \{|i|\}}{2^{b-1}-1} \approx \frac{\max_n\{\beta^n + \lambda \times \sqrt{\gamma^n}\}}{2^{b-1}-1} = s_{I_l}^{\text{static}} \in \mathbb{R}
\end{equation}
where $\lambda$ is a sensitivity parameter. This quantization method requires no additional computations at inference but only introduces a very coarse, per-layer scaling factor $s_{I_l}^{\text{static}}$.
\paragraph{Dynamic Input Quantization:} The goal is to compute $s_{I_l}^{\text{dynamic}}\in\mathbb{R}$ based on the inferred input $I_l$ at the cost of overhead computations at inference. Consequently
\begin{equation}\label{eq:dynamic}
    \frac{\max_{i\in I_l} \{|i|\}}{2^{b-1}-1} = s_{I_l}^{\text{dynamic}} \in \mathbb{R}
\end{equation}
The computation of $\max{i\in I_l} \{|i|\}$ is performed at each inference which adds a significant computational overhead (see section \ref{sec:dynamic_vs_static}). However, the scaling factor $\max{i\in I_l} \{|i|\}$ is necessarily tighter than in the static case, hence a lower quantization error. Nevertheless, we argue that it is possible to design a tighter static input quantization scheme thanks to per-channel rescaling.

\subsection{Per-Channel Static Input Quantization}\label{sec:3_2}
We define the scaling vector $s_{I_l}^{\text{channel}}\in\mathbb{R}^{n_{l-1}}$ using the BN layers. Formally, 
\begin{equation}\label{eq:per-channel}
   \frac{\max_{i\in I_l^n \text{ from }\mathcal{D}} \{|i|\}}{2^{b-1}-1} \approx  \frac{\beta^n + \lambda \times \sqrt{\gamma^n}}{2^{b-1}-1} = \left(s_{I_l}^{\text{channel}}\right)^n \quad \text{and}\quad s_{I_l}^{\text{channel}} \in \mathbb{R}^{n_{l-1}}
\end{equation}
However, we are no longer able to perform the de-quantization as described in equation \ref{eq:def} because of dimensionality issues. 
Formally, the scaling vector $s_{I_l}^{\text{channel}}$ can be applied to $I_l$ but not to the activation $\left\lfloor \frac{I_l}{s_{I_l}} \right\rceil \times \left\lfloor \frac{W_l}{s_{W_l}} \right\rceil$. To tackle this limitation, we propose to decompose the quantization in two steps. First, we update $W_l$ such that it applies both the inverse of the rescaling $s_{I_l}$ to the inputs $I_l$ and the operation originally defined by $W_l$. Then we note,
\begin{equation}\label{eq:fold}
   W_l^{\text{channel}} = \text{diag}(s_{I_l}^{\text{channel}}) \times W_l  
\end{equation}
where $\text{diag}$ is the transformation of a vector in a diagonal matrix.
Second, we scale the new value $W_l^{\text{channel}}$ as a single weight tensor. Consequently, equation \ref{eq:def} becomes:
\begin{equation}\label{eq:new_def}
    f_l^q : I_l \mapsto s_{W_l^{\text{channel}}} \odot \left( \left\lfloor \frac{I_l}{s_{I_l}^{\text{channel}}} \right\rceil \times \left\lfloor \frac{W_l^{\text{channel}}}{s_{W_l^{\text{channel}}}} \right\rceil\right)
\end{equation}
In other words, the per-channel input ranges and scaling factor $s_{I_l}^{\text{channel}}$ are computed and folded within $W_l$ (equation \ref{eq:fold}). This allows us to re-scale the input $I_l$ prior to quantization only, thus circumventing the dimensionality constraint introduced in section \ref{sec:3_1}. Moreover, this allows us to reduce the error as compared to the quantization as each output channel (or neuron) becomes:
\begin{equation}\label{eq:detailed_inference}
    \left(f_l^q(I_l)\right)^n = \sum_{m=1}^{n_{l-1}} s_{W}^n \left\lfloor\frac{I_l^m}{s_{I_l}^m}\right\rceil \times \left\lfloor(2^{b-1}-1)\frac{W^{n,m}_l/s_{I_l}^m}{\max_{m}\{|W^{n,m}_l|/s_{I_l}^m\}}\right\rceil  
\end{equation}
where $s_{I_l}^m$ is the $m^{\text{th}}$ value of $s_{I_l}^{\text{channel}}$ and $W^{n,m}_l$ is value of coordinate $n,m$. Furthermore, we deduce from equation \ref{eq:per-channel} that,
\begin{equation}
    \left\|I_l - s_{I_l}^m \left\lfloor\frac{I_l^m}{s_{I_l}^m}\right\rceil\right\| \leq \left\|I_l - s_{I_l}^{\text{static}} \left\lfloor\frac{I_l^m}{s_{I_l}^{\text{static}}}\right\rceil\right\|
\end{equation}
in other words, the quantization error on the input is lower with the per-channel method. However, this method also changes the weight quantization by folding the input scales in the weight tensor $W_l$. The difference between the static and per-channel static methods lies in the denominator $\max_{m}\{|W^{n,m}_l|/s_{I_l}^m\}$ from equation \ref{eq:detailed_inference}. By definition, we have $s_{I_l}^m \leq s_{I_l}^{\text{static}}$ and scalar values $W^{n,m}_l$ of $W_l$ are likely to be cancelled if and only if both the $W^{n,m}_l$ and corresponding $I_l^m$ have near zero ranges. We deduce that the proposed method will provide a lower quantization error on average, \textit{i.e.}
\begin{equation}\resizebox{\hsize}{!}{%
    $
    \mathbb{E}_{I\in\mathcal{D}}\left[\left\|IW - s_{W s_{I}^{\text{channel}}} \left\lfloor \frac{I}{s_{I}^{\text{channel}}} \right\rceil \left\lfloor \frac{W s_{I}^{\text{channel}}}{s_{W s_{I}^{\text{channel}}}} \right\rceil\right\|\right] \leq \mathbb{E}_{I\in\mathcal{D}}\left[ \left\|IW - s_{I}^{\text{static}}s_{W} \left\lfloor \frac{I}{s_{I}^{\text{static}}} \right\rceil \left\lfloor \frac{W}{s_{W}} \right\rceil\right\|\right]
    $}
\end{equation}
This provides an intuition on the superior performance of the proposed SPIQ quantization scheme over the reference static approach. In what follows, we show that SPIQ also empirically outperforms the dynamic approach, which in turn allows to significantly improve over current state-of-the-art methods.
\section{Experiments}
\subsection{Datasets and Implementation Details}
We validate the proposed method on three challenging computer vision tasks. First, on image classification, we consider ImageNet \cite{imagenet_cvpr09} ($\approx 1.2$M images train/50k test).
Second, on object detection, we conduct the experiments on Pascal VOC 2012 \cite{pascal-voc-2012} ($\approx$ 17k images in the test set).
Third, on image segmentation, we use the CityScapes dataset \cite{cordts2016cityscapes} (with 500 validation images).

In our experiments we tackle the challenging compression of MobileNets \cite{sandler2018MobileNetV2}, ResNets \cite{he2016deep}, EfficientNets \cite{tan2019efficientnet} and DenseNets \cite{huang2017densely} on ImageNet.
For Pascal VOC object detection challenge we test SPIQ on an SSD \cite{liu2016ssd} architecture with MobileNet backbone.
On CityScapes we use DeepLab V3+ \cite{chen2018encoder} with a MobileNet backbone.

ResNet, DenseNet, MobileNet and EfficientNet for ImageNet come from Tensorflow models zoo \cite{abadi2016tensorflow}.
In object detection, we tested the SSD model with a MobileNet backbone from \cite{manish_github}.
Finally, in image semantic segmentation, the DeepLab V3+ model came from \cite{bonlime_github}.
The networks pre-trained weights provide standard baseline accuracies on each task.
SPIQ and quantization baselines are implemented using Numpy.
The results were obtained using an Intel(R) Core(TM) i9-9900K CPU @ 3.60GHz and RTX 3090 GPU.

We performed hyper parameter settings as well as comparisons using the standard quantization operator over weight values from \cite{krishnamoorthi2018quantizing}. This operator is also applied in other recent quantization methods such as OCS \cite{zhao2019improving} and SQNR \cite{meller2019same}. For our comparison with state-of-the-art approaches in data-free quantization, we applied the more complex quantization operator from SQuant \cite{squant2022} using our own implementation which was carefully implemented so as to match the results for the original paper.

\subsection{Hyper-Parameter Setting}
\begin{figure}[!t]
    \centering
    \includegraphics[width = \linewidth]{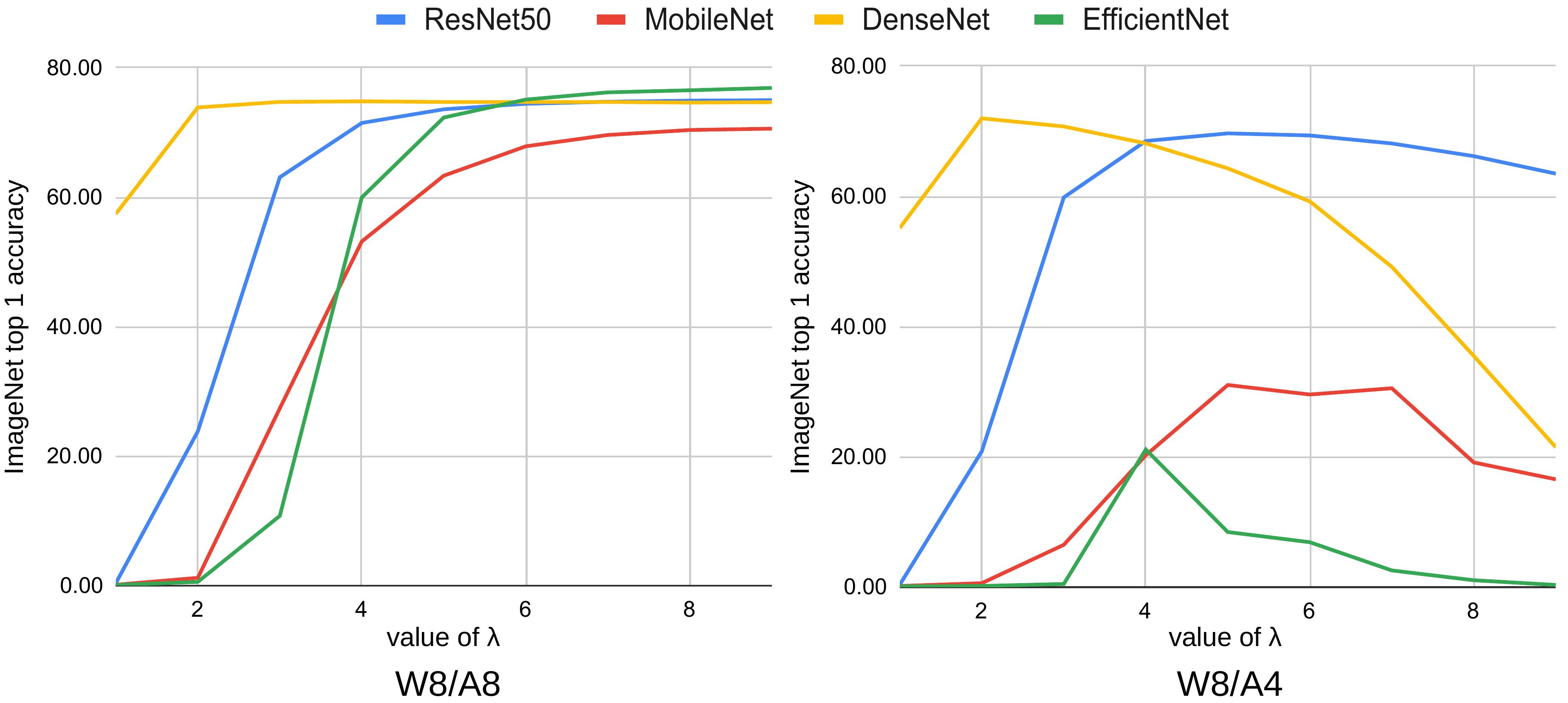}
    \caption{Influence of hyper-parameter $\lambda$ on top1 accuracy for weights quantized in int8 using the naive per-channel quantization and inputs quantized either in int8 or int4 our protocol for input quantization, on ResNet 50, MobileNet V2, DenseNet 121 and EfficientNet B0 for classification on ImageNet.}
    \label{fig:parameter_lambda}
\end{figure}

The proposed method only requires one hyper-parameter $\lambda$ which sets the number of standard deviations in the scaling value of the inputs, as defined in equation \ref{eq:per-channel}. In DFQ \cite{nagel2019data}, authors recommend setting $\lambda = 6$ for the static input quantization, based on a Gaussian prior and the objective to keep over $99.99\%$ of the input values not clipped. Intuitively, the value of $\lambda$ determines the support of the expected input distribution. In other words, a large value $\lambda$ induces almost no outliers but many small values will be quantized in a very coarse manner. On the other hand, a small value $\lambda$ induces many input outliers that will be clipped but a fine-grained quantization of smaller inputs. We empirically validate the best value for $\lambda$ and report our results in Fig \ref{fig:parameter_lambda}.
We observe that the bit-width (int4,...) has more importance than the neural network architecture on the value of $\lambda$: the smaller the representation the lower the optimal value for $\lambda$. This is a consequence of the fact that smaller bit-width can represent less values while still needing to finely quantize small input values. For the sake of simplicity, we use a common value of $\lambda$ for all architectures and define $\lambda=b$, e.g. in int8 we use $\lambda =8$ and in int4 we use $\lambda=4$. 

\subsection{Comparison with Static and Dynamic Quantization}\label{sec:dynamic_vs_static}
\begin{figure}[!t]
    \centering
    \includegraphics[width = 0.85\linewidth]{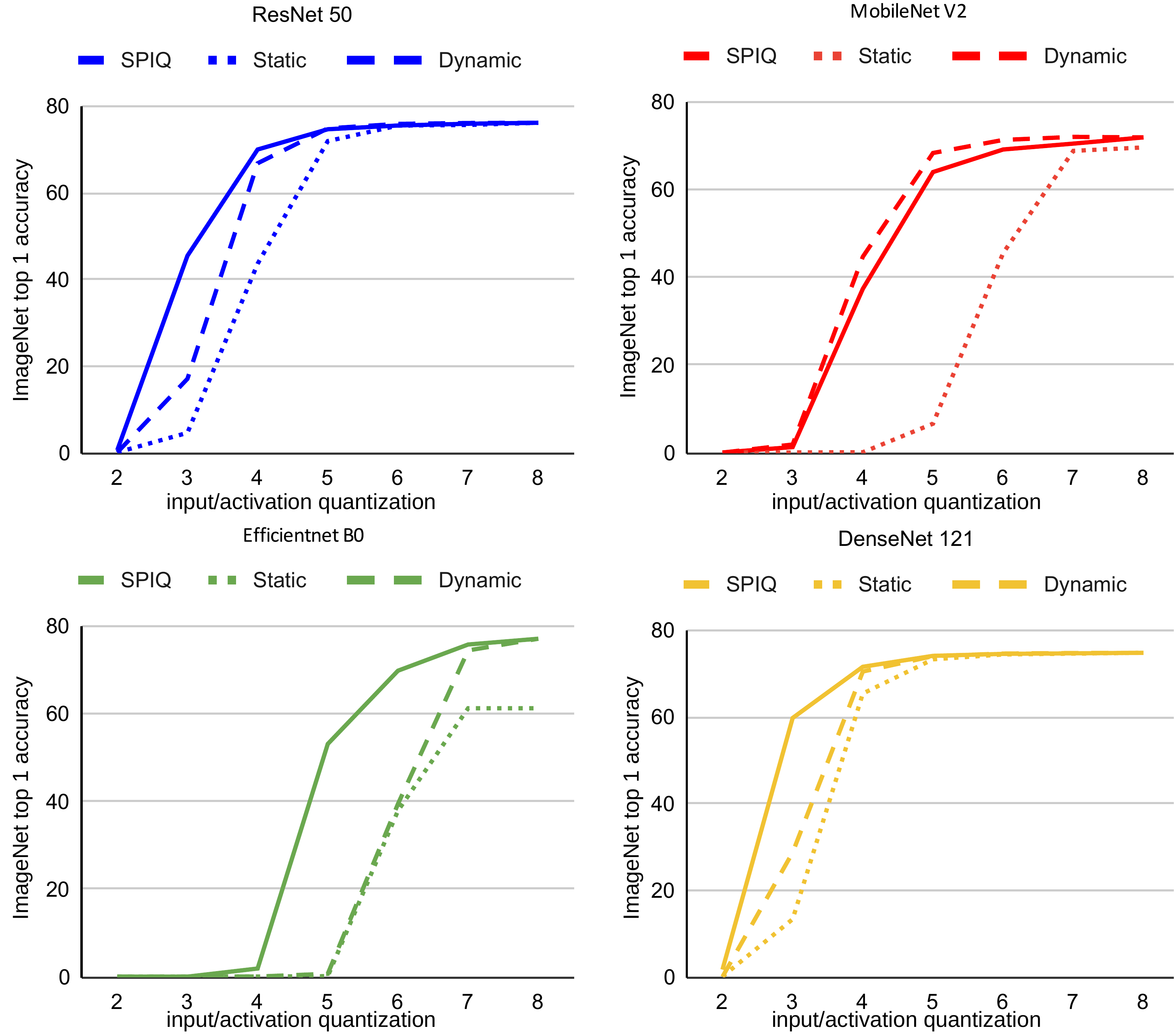}
    \caption{Comparison between SPIQ and static and dynamic inputs quantization. The weight quantization is fixed to 8 bits and we vary the input bit range from int2 (ternary quantization) to int8. We report the top1 accuracy on over ImageNet for ResNet 50, MobileNet V2, EfficientNet B0 and DenseNet 121.}
    \label{fig:comparison_dynamic_static}
\end{figure}

Fig \ref{fig:comparison_dynamic_static} presents the comparison between the static, dynamic and SPIQ method in terms of accuracy with respect to the bit-width of the inputs and activations. Given weights quantized with \cite{krishnamoorthi2018quantizing} in int8, we observe the accuracy improvement offered by the dynamic approach over the static one. For instance, on DenseNet 121 in W8/A3 (int 8 weights and int3 activations), we observe an improvement of $15.38$ points. This is due to the adaptive scaling to each input from the dynamic method. Nonetheless, the proposed per-channel manages to further improve the accuracy over the dynamic method. On the same example, SPIQ adds $46,39$ and $31,01$ points over the static and dynamic baselines respectively. The only architecture on which the dynamic and SPIQ methods achieve similar results is MobileNet V2 while on EfficientNet B0 quantized in W8/A6, SPIQ outperforms the dynamic approach by $30.35$ points. These results are a consequence of a tighter quantization for each specific channel with SPIQ.

Furthermore, in terms of inference speed, as measured in Table \ref{tab:inference}, the SPIQ method systemically outperforms the dynamic approach. For instance, on MobileNet V2 the proposed method achieves a $18\%$ faster inference. This corresponds to the cost of tuning the scaling parameters for each inputs during inference.
Consequently, SPIQ empirically offers the inference speed of the static approach and achieves accuracy results that are on par or greater than the dynamic method.

\begin{table}[!t]
\caption{Comparison of the inference time on the ImageNet validation set for different architectures quantized with the static (same runtime as SPIQ) and the dynamic methods. We report the boost induced by using the proposed static method.}
\label{tab:inference}
\centering
\setlength\tabcolsep{4pt}
\begin{tabular}{|c|c|c|c|c|c|c|}
\hline
Method & ResNet 50 & MobileNet V2 & DenseNet 121 & EfficientNet B0 \\
\hline
\hline
dynamic & 79s & 50s & 93s & 59s\\
SPIQ & 63s & 41s & 77s & 51s\\
\hline
\hline
boost & 20.2\% & 18.0\% & 17.2\% & 13.6\% \\
\hline
\end{tabular}
\end{table}

In the following section, we comapre the SPIQ performance to other data-free quantization algorithm focused on weight quantization.

\subsection{Comparison with State-Of-The-Art}
Table \ref{tab:comparison_sota_resnet} lists the performances of several data-free quantization methods on different quantization configurations of ResNet 50 on ImageNet. We classify methods by their usage of data generation (DG). Such requirement is time consuming as compared to the proposed method which takes less than a second to quantize the model while several back-propagation passes take a few minutes and fine-tuning a few hours. Nonetheless, we demonstrate that the proposed input quantization allows us to achieve superior results than other data-free quantization protocols by a large margin. Specifically, in int8 and int6 the accuracy almost reaches the full precision (float 32) accuracy while in int4, we reduce the accuracy drop by $14.56\%$ as compared to SQuant and by $68.5\%$ as compared to GDFQ \cite{xu2020generative}. This confirms that the application of the input scaling to the weights before quantization (equation \ref{eq:new_def} in section \ref{sec:3_2}) does not harm the weight quantization even in low precision. Overall, the proposed method achieves remarkable accuracy in this benchmark.

\begin{table}[!t]
\caption{Comparison between state-of-the-art, data-free, post training quantization techniques with ResNet 50 on ImageNet. We distinguish methods requiring data generation (No DG). In SPIQ the weight quantization method is SQuant.}
\label{tab:comparison_sota_resnet}
\centering
\setlength\tabcolsep{4pt}
\begin{tabular}{|c|c|c|c|c|c|}
\hline
Architecture & Method & No DG & W-bit & A-bit & Accuracy \\
\hline
\hline
\multirow{19}{*}{ResNet 50} & Baseline & - & 32 & 32 & 76.15 \\
\cline{2-6}
& DFQ \cite{nagel2019data} & \cmark & 8 & 8 & 75.45 \\
& ZeroQ \cite{cai2020zeroq} & \xmark & 8 & 8 & 75.89 \\
& DSG \cite{zhang2021diversifying} & \xmark & 8 & 8 & 75.87 \\
& GDFQ \cite{xu2020generative} & \xmark & 8 & 8 & 75.71 \\
& SQuant \cite{squant2022} & \cmark & 8 & 8 & 76.04 \\
& SPIQ & \cmark & 8 & 8 & \textbf{76.15} \\
\cline{2-6}
& DFQ \cite{nagel2019data} & \cmark & 6 & 6 & 71.36 \\
& ZeroQ \cite{cai2020zeroq} & \xmark & 6 & 6 & 72.93 \\
& DSG \cite{zhang2021diversifying} & \xmark & 6 & 6 & 74.07 \\
& GDFQ \cite{xu2020generative} & \xmark & 6 & 6 & 74.59 \\
& SQuant \cite{squant2022} & \cmark & 6 & 6 & 75.95 \\
& SPIQ & \cmark & 6 & 6 & \textbf{76.14} \\
\cline{2-6}
& DFQ \cite{nagel2019data} & \cmark & 4 & 4 & 0.10 \\
& ZeroQ \cite{cai2020zeroq} & \xmark & 4 & 4 & 7.75 \\
& DSG \cite{zhang2021diversifying} & \xmark & 4 & 4 &  23.10 \\
& GDFQ \cite{xu2020generative} & \xmark & 4 & 4 & 55.65 \\
& SQuant \cite{squant2022} & \cmark & 4 & 4 & 68.60 \\
& SPIQ & \cmark & 4 & 4 & \textbf{69.70} \\
\hline
\end{tabular}
\end{table}

\begin{table}[!t]
\caption{Comparison between state-of-the-art, data-free, post training quantization techniques with MobileNet V2, DenseNet 121 and EfficientNet B0 on ImageNet. We focused on data-free post training quantization methods that don't involve back-propagation. In SPIQ the weight quantization method is SQuant.}
\label{tab:comparison_sota_others}
\centering
\setlength\tabcolsep{4pt}
\begin{tabular}{|c|c|c|c|c|c|}
\hline
Architecture & Method & No BP & W-bit & A-bit & Accuracy \\
\hline
\multirow{7}{*}{MobileNet V2} & Baseline & - & 32 & 32 & 71.80 \\
\cline{2-6}
& DFQ \cite{nagel2019data} & \cmark & 8 & 8 & 70.92 \\
& SQuant \cite{squant2022} & \cmark & 8 & 8 & 71.68 \\
& SPIQ & \cmark & 8 & 8 & \textbf{71.79} \\
\cline{2-6}
& DFQ \cite{nagel2019data} & \cmark & 6 & 6 & 45.84 \\
& SQuant \cite{squant2022} & \cmark & 6 & 6 & 55.38 \\
& SPIQ & \cmark & 6 & 6 & \textbf{63.24} \\
\hline
\multirow{13}{*}{DenseNet 121} & Baseline & - & 32 & 32 & 75.00 \\
\cline{2-6}
& DFQ \cite{nagel2019data} & \cmark & 8 & 8 & 74.75 \\
& OCS \cite{zhao2019improving} & \cmark & 8 & 8 & 74.10 \\
& SQuant \cite{squant2022} & \cmark & 8 & 8 & 74.70 \\
& SPIQ & \cmark & 8 & 8 & \textbf{75.00} \\
\cline{2-6}
& DFQ \cite{nagel2019data} & \cmark & 6 & 6 & 73.47 \\
& OCS \cite{zhao2019improving} & \cmark & 6 & 6 & 65.80 \\
& SQuant \cite{squant2022} & \cmark & 6 & 6 & 73.62 \\
& SPIQ & \cmark & 6 & 6 & \textbf{74.54} \\
\cline{2-6}
& DFQ \cite{nagel2019data} & \cmark & 4 & 4 & 0.10 \\
& OCS \cite{zhao2019improving} & \cmark & 4 & 4 & 0.10 \\
& SQuant \cite{squant2022} & \cmark & 4 & 4 & 47.14 \\
& SPIQ & \cmark & 4 & 4 & \textbf{51.83} \\
\hline
\multirow{7}{*}{EfficientNet B0} & Baseline & - & 32 & 32 & 77.10 \\
\cline{2-6}
& DFQ \cite{nagel2019data} & \cmark & 8 & 8 & 76.89 \\
& SQuant \cite{squant2022} & \cmark & 8 & 8 & 76.93 \\
& SPIQ & \cmark & 8 & 8 & \textbf{77.02} \\
\cline{2-6}
& DFQ \cite{nagel2019data} & \cmark & 6 & 6 & 43.08 \\
& SQuant \cite{squant2022} & \cmark & 6 & 6 & 54.51 \\
& SPIQ & \cmark & 6 & 6 & \textbf{74.67} \\
\cline{2-6}
\hline
\end{tabular}
\end{table}

To further validate the efficiency of SPIQ, in Table \ref{tab:comparison_sota_others}, we report results on DenseNet 121, EfficientNet and MobileNet V2. The considered architectures, especially MobileNet V2 and EfficientNet, are even more challenging than ResNet to quantize without accuracy drop even in relatively large representations such as int6. We only focused on the state-of-the-art approaches (without data generation) OCS \cite{zhao2019improving}, DFQ \cite{nagel2019data} and SQuant \cite{squant2022}. We observe the large benefits of a stronger input quantization method as SPIQ improves by $7.86\%$ the accuracy over SQuant and $17.4\%$ over DFQ on MobileNet V2 in int6. The results are even more impressive on EfficientNet B0 in int6, as SPIQ improves the accuracy by $20.16\%$ than SQuant and $31.59\%$ over DFQ. As compared to OCS, on DenseNet 121, the proposed method boosts the accuracy by $8.74\%$.
Still, data-free quantization has room for improvement in int4 quantization, on already efficient architectures such as MobileNet V2 and EfficientNet B0.
In the following section, we propose to generalize these remarkable results to other challenging computer vision tasks.

\subsection{Other Applications}
In addition to results on classification networks, we show the interest of SPIQ for both semantic segmentation and object detection.

\paragraph{Semantic Segmentation:} In table \ref{tab:segmentation}, we report the performance of SPIQ method on image semantic segmentation task of CityScapes dataset. The dynamic approaches still provides more accuracy than the static baseline due to its adaptive scaling of each input regardless of the weight quantization process. Still, due to a finer quantization of the inputs for each channel, SPIQ manages to further improve the accuracy over the dynamic method reaching outstanding results such as $68.69$ mIoU in W6A6.
This confirms the two previous main results: first, SPIQ offers the highest accuracy while preserving the inference-time benefits of static input quantization. Second, when used in combination with a strong weight quantization protocol, SPIQ achieves state-of-the-art performances and significantly improve the accuracy in low-bit representation (int4). More precisely, we improve by $29.63\%$ the accuracy the mean intersection over union (mIoU) of a DeepLab V3+ with MobileNet backbone on CityScapes.
\begin{table}[!t]
\caption{Performance (mIoU) on semantic segmentation on CityScapes dataset.}
\label{tab:segmentation}
\centering
\setlength\tabcolsep{4pt}
\begin{tabular}{|c|c|c|c|c||c|}
\hline
Architecture & method & W4/A4 & W6/A6 & W8/A8 & W32/A32 \\
\hline
\hline
\multirow{6}{*}{DeepLab V3+} & baseline & - & - & - & 70.71 \\
& DFQ + static & 6.51 & 45.71 & 70.11 & - \\
& DFQ + dynamic & 7.51 & 66.65 & 70.22 & - \\
& SQuant + static & 7.69 & 66.77 & 70.21 & - \\
& SQuant + dynamic & 28.87 & 66.98 & 70.42 & - \\
& SPIQ & \textbf{36.14} & \textbf{68.69} & \textbf{70.66} & - \\
\hline
\end{tabular}
\end{table}

\paragraph{Object Detection:} In Table \ref{tab:detection}, we report the performance of SPIQ method on object detection of Pascal VOC 2012 dataset. Dynamic input quantization outperforms the static baseline in terms of accuracy at the expense of runtime. Nonetheless, SPIQ manages to further improves the mAP by $2.41$ points. This is a consequence of the fine-grained quantization suited for each input channel of each layers of the network, in all bit-width configurations.
These results confirm our two main results: SPIQ offers the highest mean average precision (mAP) in all quantization configurations as compared to static and dynamic methods, from int8 to low-bit int4. Furthermore, the SPIQ method achieves higher mAP than other state-of-the-art quantization schemes that focus only on improving weight quantization. These results conclude our empirical validation of the great performance of the proposed method.
\begin{table}[!t]
\caption{Performance (mAP) on object detection on Pascal VOC 2012 dataset.}
\label{tab:detection}
\centering
\setlength\tabcolsep{4pt}
\begin{tabular}{|c|c|c|c|c||c|}
\hline
Architecture & method & W4/A4 & W6/A6 & W8/A8 & W32/A32 \\
\hline
\hline
\multirow{6}{*}{SSD MobileNet} & baseline & - & - & - & 68.56 \\
& DFQ + static & 3.94 & 53.52 & 67.91 & - \\
& DFQ + dynamic & 15.95 & 62.31 & 67.52 & - \\
& SQuant + static & 14.98 & 61.29 & 68.43 & - \\
& SQuant + dynamic & 35.47 & 66.72 & \textbf{68.56} & - \\
& SPIQ & \textbf{37.88} & \textbf{68.01} & \textbf{68.56} & - \\
\hline
\end{tabular}
\end{table}

\section{Discussion}
\begin{figure}[!t]
    \centering
    \includegraphics[width = \linewidth]{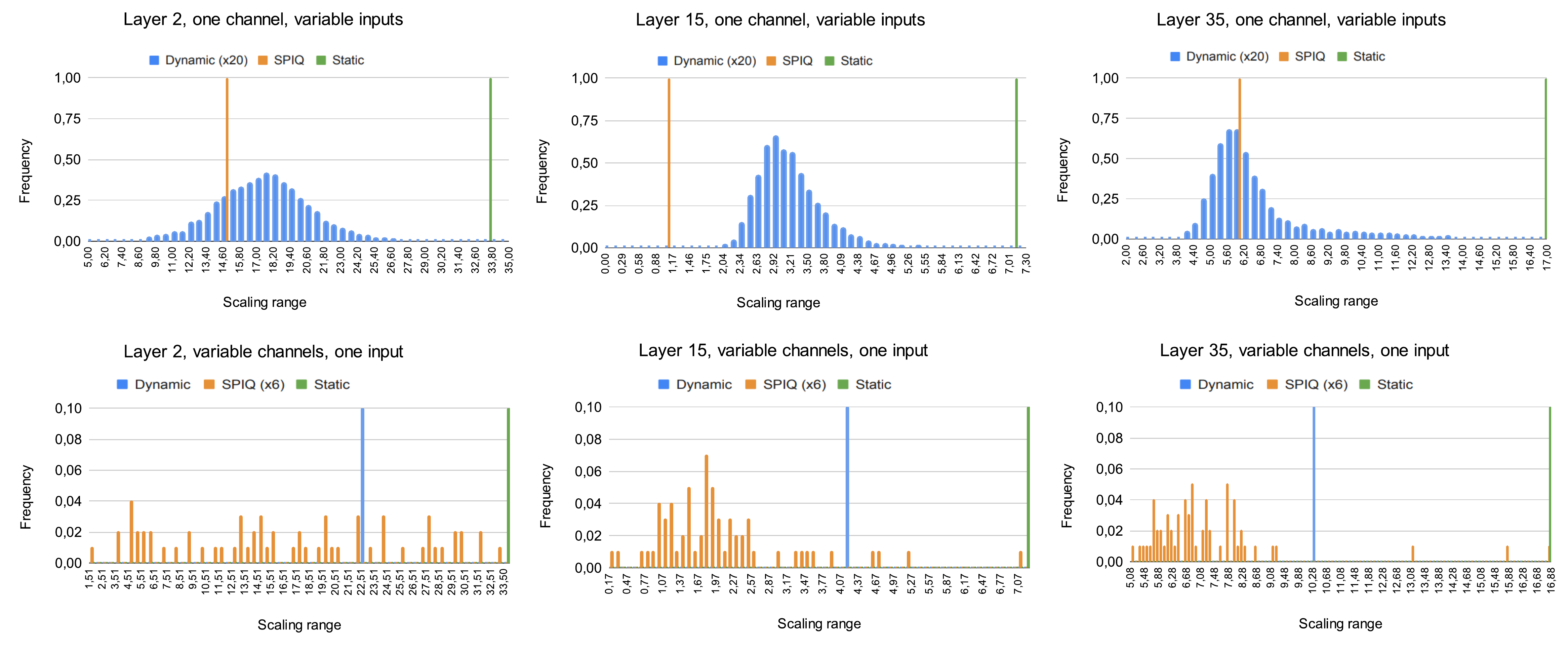}
    \caption{Distribution of the quantization ranges outputted by SPIQ, the dynamic and static baselines on the inputs of 3 different layers of a ResNet 50. The static baseline is constant, while dynamic and SPIQ vary depending on the input samples and channels respectively. The lower the computed ranges to 0 (the closer to the left of the each subplot), the better. SPIQ generally allows tighter adaptation to the original input distribution, as compared with both the static and dynamic methods.}
    \label{fig:qualitative_analysis}
\end{figure}

\begin{figure}[!t]
    \centering
    \includegraphics[width = 0.8\linewidth]{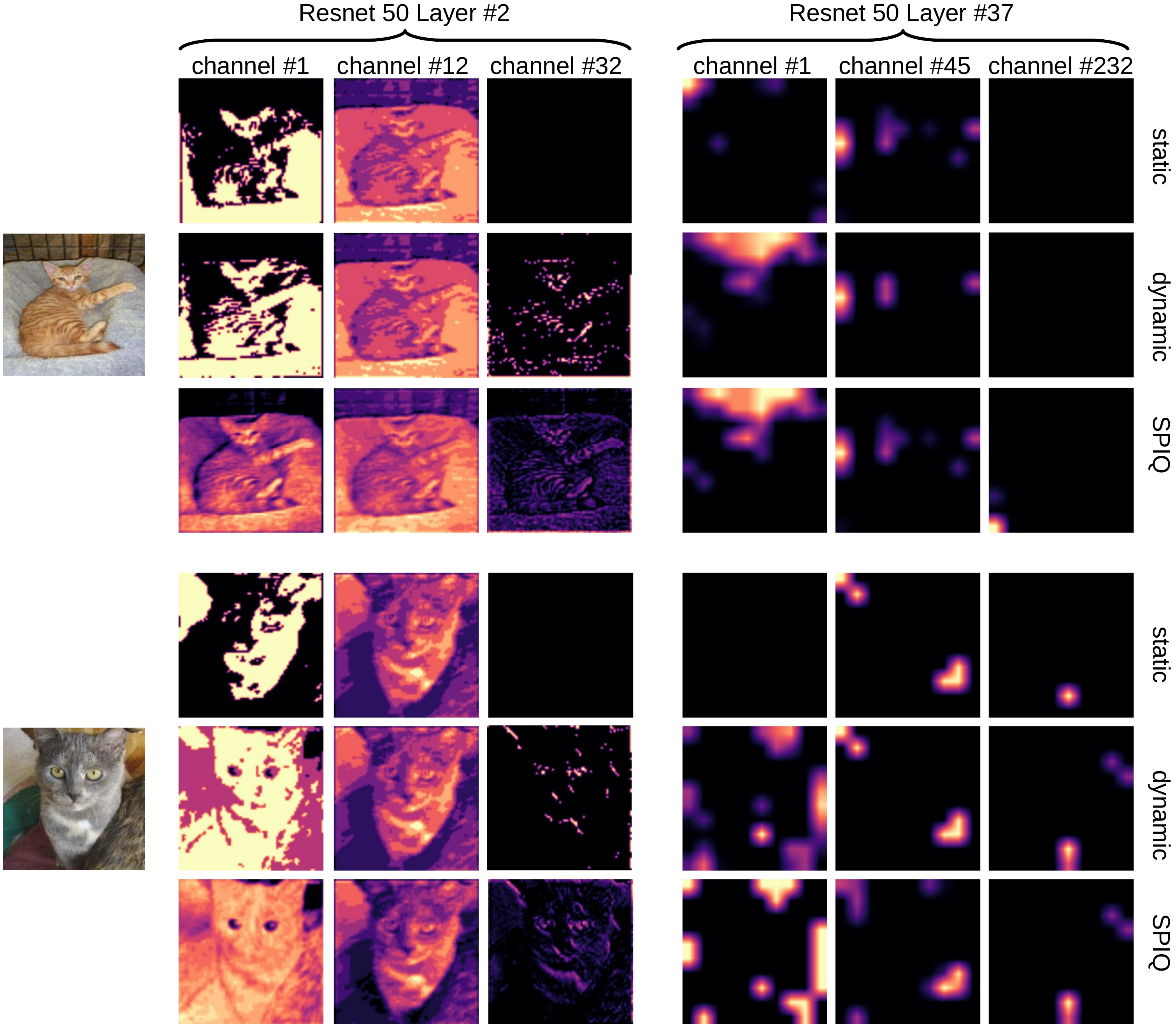}
    \caption{Illustration of some channels of different feature maps of a ResNet 50 quantized with the static, dynamic and proposed SPIQ method. We observe sharper spatial features when using the proposed method.}
    \label{fig:feature_maps}
\end{figure}

\paragraph{Empirical Intuition:} Fig \ref{fig:qualitative_analysis} shows a comparison of sample scaling ranges calculated with the static and dynamic approaches as well as SPIQ. It stems from the definition of these methods that the closer the range is to 0 (on the left on Fig \ref{fig:qualitative_analysis} subplots), the tighter the quantized inputs to the original inputs. Furthermore, while the static range is the same across all examples and channels, the dynamic method as well as SPIQ respectively vary upon those two factors. We observe that the static approach is not very tight to the input distribution in all cases. The dynamic approach, allows tighter adaptation in every scenario. Furthermore, depending on the input example (first row of Fig \ref{fig:qualitative_analysis}), SPIQ is generally tighter than the dynamic approach (most notably on e.g. layer 15 where the range computed with SPIQ is far lower, and to a lesser extent on e.g. layer 2). Furthermore, varying the input channels with one fixed example (second row of Fig \ref{fig:qualitative_analysis}) shows that the ranges computed with SPIQ are generally tighter than those computed with the dynamic approach. Fig \ref{fig:feature_maps} also illustrates how, on certain channels (e.g. channel 32 of layer 2), the dynamic approach struggles to leverage the full quantized range of values. By contrast, SPIQ qualitatively allows to better preserve feature map details, which in turn improves the accuracy. Hence, we argue that, if one had to chose between per-example and per-channel quantization, the latter would be more relevant. However, why wouldn't we do both?


\paragraph{On the possibility to design a Per-Channel Dynamic Quantization:} Per-channel dynamic quantization could mathematically be performed by simply combining equation \ref{eq:dynamic} and equation \ref{eq:per-channel}. However, in practice this would require to perform weight quantization in addition to activations quantization at each inference step. This would be extremely time-consuming, especially when dealing with fully-connected layers that have larger weight tensors than input tensors. Furthermore, this would require to store weight values in full precision instead of low-bit precision which removes one of the benefits of quantization, \textit{i.e.} memory foot-print reduction. Consequently, while per-channel dynamic quantization is theoretically feasible, in practice one has to choose between per-example and per-channel modelling as combining the two is highly impractical. As such, we show that SPIQ allows to better preserve the channel-wise information, leading to improved accuracies for the quantized networks with lower inference runtimes.
\section{Conclusion}
In this work, we highlighted a current limitation of post-training quantization methods, arguing that quantizing the inputs of each layer is of paramount importance to successful PTQ, that is often neglected in the literature. Furthermore, we showed that per-channel
range estimation allows tighter modelling of the full-precision distribution e.g. as compared to per-example, dynamic approaches. Thus, we proposed SPIQ, a novel static input quantization approach which leverages per-channel quantization of the inputs in a data-free manner. We empirically showed that SPIQ achieved better speed vs. accuracy trade-offs than both the static and dynamic input methods, in addition to significantly outperform existing state-of-the-art methods across a wide range of applications and neural network architectures without bells and whistles.

\paragraph{Limitations and Future Work:}
Very low-bit representation remains an extremely challenging task for data-free acceleration. In cases such as binary or ternary quantization, the proposed method would greatly benefit from fine-tuning. Generated data, obtained with similar methods as \cite{zhang2021diversifying,xu2020generative}, may provide better insight on inputs and activation distributions, which could in theory improve the scales estimation for input quantization.

\bibliographystyle{splncs04}
\bibliography{egbib}
\clearpage

\end{document}